\documentclass[acmtog]{acmart}

\usepackage{datenumber}
\usepackage{calc}
\usepackage[mmddyyyy]{datetime}
\newcounter{datetoday}
\newcounter{diffyears}
\newcounter{diffmonths}
\newcounter{diffdays}
\newcommand{\difftoday}[3]{%
      \setmydatenumber{datetoday}{\the\year}{\the\month}{\the\day}%
      \setmydatenumber{diffdays}{#1}{#2}{#3}%
      \addtocounter{diffdays}{-\thedatetoday}%
      \ifnum\value{diffdays}>0
        \def\diffbefore{}%
        \def\diffafter{left}%
      \else
        \def\diffbefore{}%
        \def\diffafter{ago}%
        \setcounter{diffdays}{-\value{diffdays}}%
      \fi
      \setcounter{diffyears}{\value{diffdays}/365}%
      \setcounter{diffdays}{\value{diffdays}-365*\value{diffyears}}%
      \setcounter{diffmonths}{\value{diffdays}/30}%
      \setcounter{diffdays}{\value{diffdays}-30*\value{diffmonths}}%
      \diffbefore
      \ifnum\value{diffyears}=0
      \else
        \ifnum\value{diffyears}>1
            \thediffyears\space years,
        \else
            \thediffyears\space year,
        \fi
      \fi
      \ifnum\value{diffmonths}=0
      \else
        \ifnum\value{diffmonths}>1
            \thediffmonths\space months
        \else
            \thediffmonths\space month
        \fi
      \fi
      \ifnum\value{diffdays}=0
      \else
        \ifnum\value{diffdays}>1
            \thediffdays\space days
        \else
            \thediffdays\space day
        \fi
      \fi
      \diffafter
}

\newcommand{\NEW}[1]{{{{{#1}}}}}

\setcitestyle{square}

\usepackage{amsbsy}
\usepackage[mathscr]{euscript}
\usepackage{amsmath}
\usepackage{stackrel}
\usepackage{nicefrac}
\usepackage{kotex}
\usepackage{algpseudocode}
\usepackage[ruled]{algorithm2e} 

\SetAlFnt{\small}
\SetAlCapFnt{\small}
\SetAlCapNameFnt{\small}
\SetAlCapHSkip{0pt}
\IncMargin{-\parindent}

\usepackage{dblfloatfix}
\usepackage{float}

\usepackage{subcaption}

\copyrightyear{2026}
\acmYear{2026}
\setcopyright{cc}
\setcctype{by}
\acmConference[SIGGRAPH Conference Papers '26]{Special Interest Group on Computer Graphics and Interactive Techniques Conference Conference Papers}{July 19--23, 2026}{Los Angeles, CA, USA}
\acmBooktitle{Special Interest Group on Computer Graphics and Interactive Techniques Conference Conference Papers (SIGGRAPH Conference Papers '26), July 19--23, 2026, Los Angeles, CA, USA}
\acmDOI{10.1145/3799902.3811078}
\acmISBN{979-8-4007-2554-8/2026/07}


\begin{document}
\title{Ambient-robust Inverse Rendering using \NEW{Active} RGB-NIR Imaging}

\author{Hoon-Gyu Chung}
\authornote{Both authors contributed equally to this research.}
\email{hgchung@postech.ac.kr}
\affiliation{%
  \institution{POSTECH}
  \country{Republic of Korea}
}

\author{Jinnyeong Kim}
\authornotemark[1]
\email{wlssud0701@postech.ac.kr}
\affiliation{%
  \institution{POSTECH}
  \country{Republic of Korea}
}

\author{Hyunwoo Kang}
\email{cein39296@postech.ac.kr}
\affiliation{%
  \institution{POSTECH}
  \country{Republic of Korea}
}

\author{Seung-Hwan Baek}
\email{shwbaek@postech.ac.kr}
\affiliation{%
  \institution{POSTECH}
  \country{Republic of Korea}
}

\begin{abstract}
Inverse rendering aims to reconstruct geometry and reflectance of objects from images.
Despite recent progress, existing methods often produces inaccurate reconstructions that are sensitive to ambient illumination conditions.
Here we introduce an ambient-robust inverse rendering method enabled by \NEW{active} RGB–NIR imaging. Our key insight is to leverage near-infrared (NIR) flash illumination—imperceptible to human observers—to obtain stable point-light shading that is largely invariant to ambient illumination.
By using multi-view RGB images illuminated by ambient light and NIR images acquired with active NIR flash illumination, we reconstruct accurate geometry and reflectance by exploiting the complementary benefits of RGB and NIR images via a three-stage inverse rendering method.
To enable dense multi-view acquisition, we develop an active imaging system equipped with a RGB–NIR camera and a NIR flash mounted on a mobile base. Using this system, we collect the first multi-view RGB–NIR inverse rendering dataset captured under multiple ambient illumination conditions.
Experiments demonstrate that our method outperforms prior approaches, achieving accurate geometry and reflectance estimation across multiple ambient lighting scenarios.
\end{abstract}

\begin{teaserfigure}
    \includegraphics[width=\linewidth]{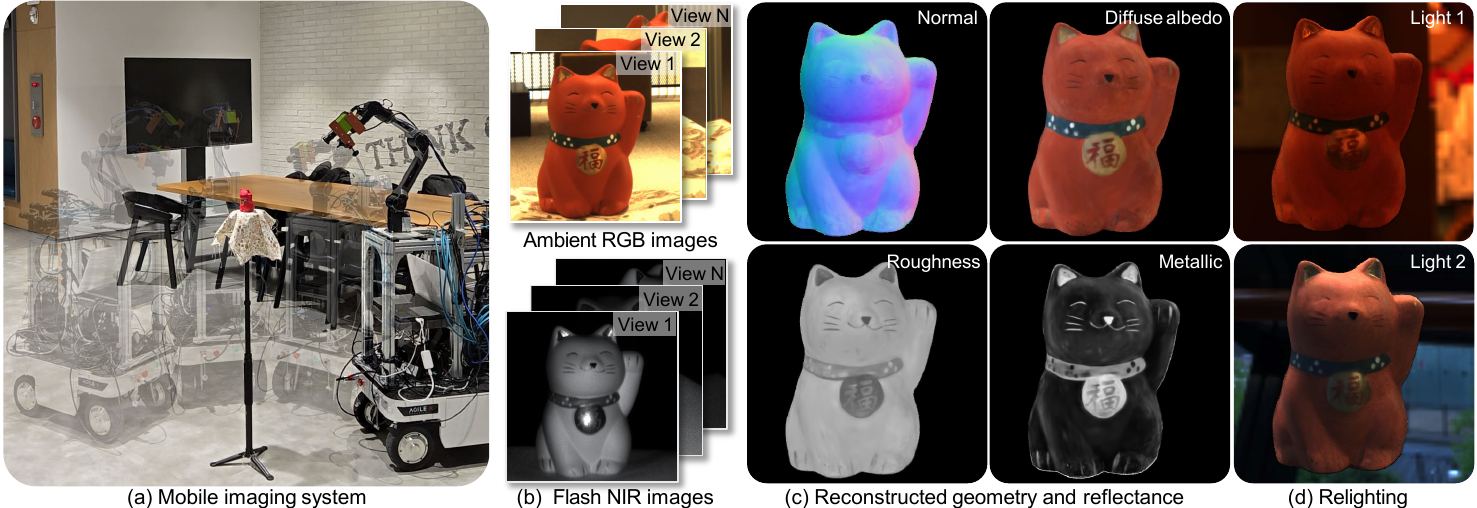}
    \captionof{figure}{
(a) We present an imaging system comprised  of an RGB–NIR camera and an NIR flash mounted on a robotic arm attached to a mobile base. Using this system, we automatically capture (b) multi-view RGB–NIR images of an object under active NIR illumination.
From these RGB--NIR observations, we demonstrate accurate inverse rendering of (c) geometry and reflectance in a manner robust to ambient lighting, which enables (d) natural relighting.
}
    \label{fig:teaser}
\end{teaserfigure}

\maketitle

\section{Introduction}
Reconstructing geometry and reflectance of objects from images —known as \emph{inverse rendering}—is a central problem in graphics and vision. Despite recent progress, applying inverse rendering under uncontrolled ambient illumination remains challenging. Ambient lighting introduces strong ambiguities between reflectance and illumination, leading to inconsistent and inaccurate reconstructions, particularly for passive methods that do not employ active illumination~\cite{nerfactor, tensoir, R3DG2023, jiang2024gaussianshader, gu2024IRGS, wei2020object, DiffusionRenderer}.
While active illumination can alleviate this ambiguity~\cite{cheng2023wildlight, zeng2023nrhints, bi2024rgs, chung2024differentiable}, using a strong light source that overwhelms ambient light in the visible spectrum can easily disturb people and environment, making such approaches impractical for in-the-wild scenarios outside darkroom settings.

This paper addresses these limitations by proposing an ambient-robust inverse rendering method using \NEW{active} RGB–NIR imaging. Our key insight is to exploit the complementary roles of RGB and NIR light. Since NIR light is imperceptible to human observers, we can employ a strong NIR flash without disturbing people and environment. This provides point-light shading measurements that remain stable under diverse ambient lighting conditions, enabling robust estimation of geometry and NIR reflectance. Simultaneously, we capture RGB images that remain naturally illuminated by the environment. 
The RGB images are used to obtain visible-spectrum reflectance, which is useful for most downstream applications. 
This is feasible as our NIR flash emits light beyond the spectral coverage of the RGB sensor.

To leverage the complementary properties of the RGB and NIR images, we introduce a three-stage RGB–NIR inverse rendering method. First, we initialize geometry using multi-view RGB images, which is robust to uncontrolled lighting. Second, we use multi-view NIR flash images to estimate surface roughness and metallic parameters, while refining geometry using reliable point-light shading that is robust to ambient illumination. Third, given accurate geometry and NIR reflectance, we recover RGB diffuse albedo and RGB environment map with our RGB--NIR BRDF model that couples NIR and RGB reflectance behavior. This three-stage RGB–NIR inverse rendering yields consistent and accurate geometry and reflectance across varied ambient lighting conditions.

To acquire dense multi-view RGB--NIR observations in real-world environments, we develop an \NEW{active} imaging system composed of a four-wheeled base and a robotic arm equipped with a pixel-aligned RGB–NIR camera and a synchronized NIR flash. 
The system autonomously scans objects by coordinating the base, arm, camera, and flash, enabling dense multi-view capture. Using this platform, we collect the first multi-view RGB–NIR inverse rendering dataset containing RGB–NIR images with flash and no-flash NIR channels under multiple ambient illumination conditions.
Experiments on both real and synthetic datasets demonstrate that our method outperforms existing approaches, achieving accurate estimation of geometry and reflectance.

In summary, our contributions are:
\begin{itemize}
    \item Ambient-robust three-stage RGB–NIR inverse rendering framework that uses multi-view RGB–NIR images with NIR flash to recover geometry and reflectance.
    \item Automated \NEW{active} RGB–NIR imaging system that captures dense, object-centric multi-view data without disturbing people and scene.
    \item RGB–NIR inverse rendering dataset consisting of multi-view RGB–NIR images with NIR flash captured across multiple ambient lighting conditions.
\end{itemize}
\section{Related Work}
\label{sec:related}

\subsection{Inverse Rendering using RGB Images}
Inverse rendering using RGB images has been extensively studied~\cite{barron2014shape, dong2014appearance, shape, mvps, efficient, practical, xia2016recovering, mvpsiso}. Recent learning-based approaches leverage CNNs~\cite{sengupta2019neural, sang2020single, li2018learning, wei2020object}, transformers~\cite{zhu2022irisformer, ikehata2023sdmunips}, and diffusion models~\cite{chen2024intrinsicanything, litman2025materialfusion, DiffusionRenderer, neural_lightrig} to estimate geometry and reflectance using learned priors. 
Analysis-by-synthesis methods invert the rendering equation through differentiable rendering frameworks, including volumetric rendering~\cite{nerfactor, tensoir, physg, invrender, iron, wu2023nefii, wang2024inverse, fei2024vminer, zeng2023nrhints, cheng2023wildlight}, 3D Gaussian splatting~\cite{R3DG2023, liang2024gs, jiang2024gaussianshader, zhu2024gs, bi2024rgs}, and 2D Gaussian splatting~\cite{gu2024IRGS, yao2025refGS}. These approaches rely on RGB images captured under fixed environmental illumination. Consequently, they struggle with the inherent ambiguity between illumination and reflectance, remaining highly sensitive to variations in ambient lighting.
\NEW{Polarimetric inverse rendering introduces additional constraints through polarization cues, but remains limited under complex illumination~\cite{dave2022pandora, Li_NeISF_CVPR2024}.}
Active illumination techniques mitigate this ambiguity by introducing controlled lighting~\cite{cheng2023wildlight, practical, zeng2023nrhints, bi2024rgs, chung2024differentiable, chung2025differentiable}. However, visible-light flash systems typically require dark-room conditions to isolate lighting effects, and their strong visible active illuminations make them impractical in everyday environments, disturbing people and environments. 
In contrast, our RGB–NIR inverse rendering leverages invisible NIR flash illumination and RGB–NIR imaging to achieve accurate inverse rendering under diverse ambient lighting conditions.

\subsection{NIR Imaging for Scene Reconstruction}
NIR imaging has been widely used in perception often together with RGB imaging, where invisible NIR illumination provides robust vision cues for textureless and dark scenes. 
NIR photometric stereo demonstrates that NIR reflectance exhibits more uniform albedo characteristics than that of visible spectrum, enabling improved geometry estimation~\cite{choe2017refining, choe2014exploiting, yoon2016fine, choe2016simultaneous}.
Using NIR and RGB images has shown to be effective in stereo depth estimation especially in dark scenes~\cite{carfagni2019metrological, kim2024pixel, brucker2024cross}.
In the context of inverse rendering, there are few works adopting NIR imaging which use a Kinect sensor equipped with structured NIR illumination, NIR camera, and an RGB camera~\cite{wu2015appfusion, ha2020progressive}.
While the benefits of invisible NIR light holds, the high-frequency structured NIR illumination in Kinect introduces reconstruction artifacts.
Also, they rely on voxel-based scene representations, which limit inverse rendering accuracy and fine-scale reflectance recovery.
We achieve accurate inverse rendering by using RGB–NIR image pairs with unstructured NIR flash, and our three-stage inverse rendering using Gaussian primitive. 


\begin{figure*}[t]
\includegraphics[width=\linewidth]{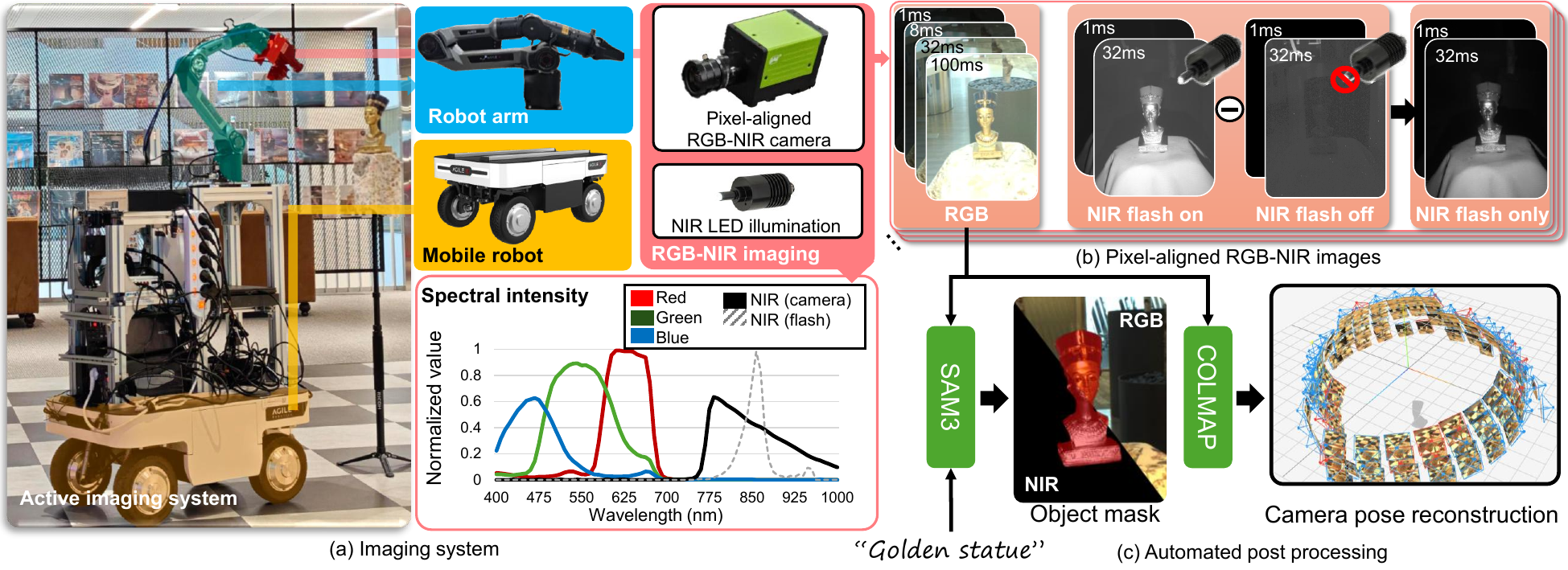}
\caption{\textbf{\NEW{RGB--NIR} vision system and image processing pipeline.} 
(a) Our vision system is composed of a robotic arm, a mobile robot base, and a pixel-aligned RGB–NIR camera with an NIR flash. Insets show each component as well as the spectral profile of the camera and the NIR flash.
(b) Sample of RGB-NIR image pairs. We acquire RGB and NIR frames with multiple exposure time. For NIR, we subtract NIR flash on image and NIR flash off image to remove ambient contributions, producing an NIR flash-only image.
(c) Image processing pipeline including mask extraction using SAM3~\cite{carion2025sam3segmentconcepts} and camera-pose reconstruction using COLMAP~\cite{Schonberger_Frahm_2016_colmap}.
}
\vspace{-4mm}
\label{fig:imaging_real}
\end{figure*}

\subsection{Imaging Systems for Inverse Rendering}
Accurate inverse rendering often relies on manual capture, either through handheld camera motion~\cite{tensoir, invrender, zeng2023nrhints, bi2024rgs} or repeated tripod repositioning~\cite{kuang2023stanfordorb, Ummenhofer2024OWL}, making acquisition labor-intensive. 
Light-stage systems do not require manual capture interventions. However, they require a fixed large-scale infrastructures~\cite{liu2024openillumination, chabert2006relighting, debevec2000acquiring}. Display-based illumination offers a practical alternative, however suffers from low intensity and narrow angular coverage~\cite{choi2025real,choi2024differentiable}.
Robotic imaging systems have emerged as a solution to these challenges because robotic arms equipped with cameras can provide programmable, repeatable viewpoints and built-in calibration.
However, previous robotic inverse-rendering systems are not mobile, restricting the range of achievable viewpoints and point-light sources~\cite{Toschi_2023_CVPR, kumar2025mobile}.
Our \NEW{active} imaging system is mobile, addressing these limitations by enabling flexible object-centric scanning with large viewpoint diversity and synchronized RGB–NIR data capture. The resulting dataset contains many views, precise camera parameters, and paired flash/no-flash NIR images and RGB images, making it suited for benchmarking inverse rendering under multiple ambient illuminations.
\section{RGB--NIR Vision System}
To automatically acquire the data required for RGB--NIR inverse rendering, we develop a vision platform shown in Figure~\ref{fig:imaging_real}(a).  
The imaging head consists of a RGB--NIR camera (JAI FS-1600) and a synchronized NIR flash (Advanced Illumination AL295).  
Using the NIR flash is beneficial compared to the RGB counterpart as the effective dynamic range can be more broaden without disturbing human observers (See the Supplemental Document for further details). 
The RGB-NIR camera employs a dichroic prism to split visible and NIR wavelengths onto dedicated RGB and NIR CMOS sensors, enabling simultaneous capture of pixel-aligned RGB and NIR frames.  
The spectral sensitivity functions of the two sensors and the emission spectrum of the NIR flash are shown in Figure~\ref{fig:imaging_real}(a).
The imaging head is mounted on a robotic arm (AgileX Piper) using a custom 3D-printed holder, and the arm is rigidly attached to a wheeled mobile base (AgileX Ranger Mini V2), forming a fully mobile acquisition platform.

\begin{figure}[t]

\includegraphics[width=\linewidth]{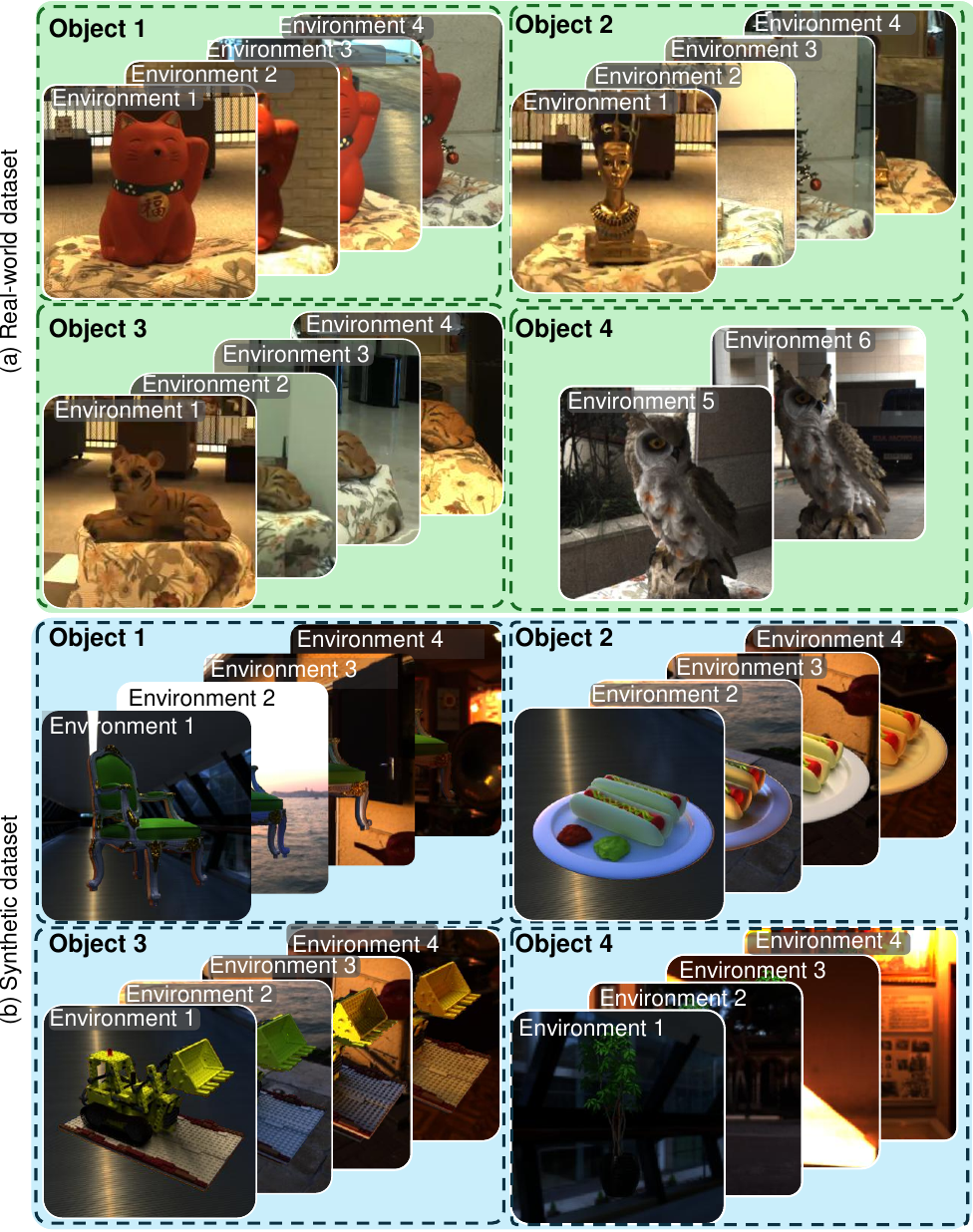}
\caption{\NEW{\textbf{RGB-NIR inverse rendering dataset.} {(a) We capture multi-view RGB–NIR image pairs with active NIR flash for four real-world objects under different ambient illumination conditions.
(b) We render synthetic dataset using Mitsuba 3~\cite{mitsuba3}, matching the real-world acquisition setup.}}
}
\vspace{-12mm}
\label{fig:dataset_real}
\end{figure}

\begin{figure*}[t]
\centering
\includegraphics[width=\linewidth]{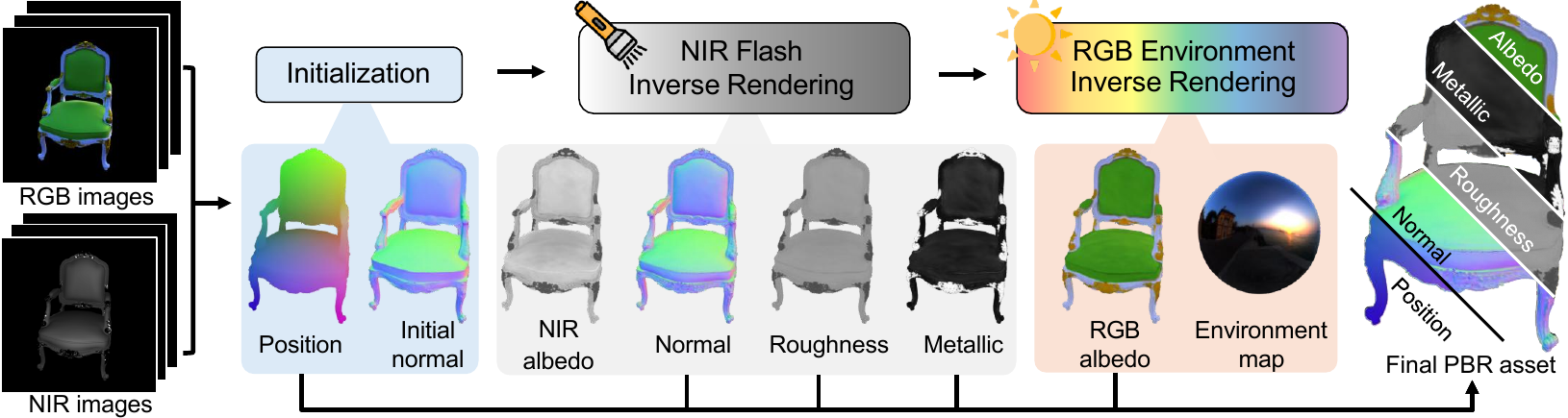}
\caption{\textbf{Three-stage RGB--NIR inverse rendering.}  
We initialize geometry using RGB images. NIR images with flash lighting are then used to estimate reflectance in a manner robust to ambient lighting and further refine geometry. RGB albedo and RGB environment map are then estimated using RGB images. 
}
\vspace{-3mm}
\label{fig:pipeline}
\end{figure*}

\paragraph{Automatic Acquisition and Image Processing}
Using the RGB--NIR vision system, we automatically collect dense multi-view observations around each object, taking around 20 minutes. 
\NEW{A single HDR RGB-NIR image capture takes roughly 12 seconds during the capture process.} See Supplemental Video for the actual capture process.
The mobile base follows a circular trajectory around the target, stopping at predefined azimuth angles.  
At each stop, the robotic arm introduces controlled out-of-plane variation by raising the imaging head through several discrete elevations.  
At every elevation, the system captures HDR RGB and HDR NIR images—with and without the NIR flash—using sensor-specific exposure bracketing.
After acquisition, the passive and active NIR images are used to isolate the contribution of the flash.  
Subtracting image captured without NIR flash $I^\text{NIR-off}$ from the image captured with NIR flash $I^\text{NIR-on}$ yields the flash-only NIR measurement $I^\text{NIR}$:
$I^\text{NIR}=I^\text{NIR-on}-I^\text{NIR-off}$
as shown in Figure~\ref{fig:imaging_real}(b).
We then generate object masks using SAM3~\cite{carion2025sam3segmentconcepts}, which enables language–driven segmentation on RGB images. 
For each dataset, we design a set of textual prompts that explicitly refer to the target object category, allowing SAM3 to suppress background regions and consistently extract only the foreground objects of interest. 
The entire sequence—including navigation, arm positioning, exposure control, flash synchronization, and image—is automated.  
We estimate camera poses using structure-from-motion~\cite{Schonberger_Frahm_2016_colmap}. 
Please refer to the Supplemental Document for the details of the image acquisition and image processing.

\section{RGB-NIR Inverse Rendering Dataset}
\paragraph{Real Scenes}
Using automated acquisition, we create a real-world RGB--NIR inverse-rendering dataset, shown in Figure~\ref{fig:dataset_real}(a).  
\NEW{We capture three objects across four indoor environments and one object across two outdoor environments.}
For each scene, we first place a mirror ball on an object holder to obtain a ground-truth environment map for the purpose of evaluation, following the procedure of Debevec et al.~\shortcite{debevec2008rendering}. We then replace the mirror ball with target objects one by one, and the robot platform executes a full acquisition sweep.  
Each object--environment pair contains over 100 synchronized RGB, ambient NIR, and flash-illuminated NIR images along with object masks and camera poses.  

\paragraph{Synthetic Scenes}
In addition to the real-world scenes, we create a synthetic RGB--NIR dataset.  
\NEW{Four objects are rendered in four different environments as shown in Figure~\ref{fig:dataset_real}(b).}  
For each object part, we manually assign reflectance parameters from the visible-to-NIR BRDF dataset of Dupuy et al.~\shortcite{dupuy2018adaptive}.  
Using Mitsuba~3~\cite{mitsuba3}, we then render RGB-NIR images under environment maps~\cite{tensoir} and an NIR point light. 
Further details of the synthetic data generation process are provided in the Supplemental Document.
\section{Three-stage RGB--NIR Inverse Rendering}
\label{sec:inverse_rendering}

We perform inverse rendering in three stages for ambient robustness:
(1) We initialize geometry using multi-view RGB images.  
(2) The geometry is refined and NIR reflectance is estimated using multi-view NIR images.  
(3) The RGB diffuse albedo and environment illumination are recovered from the multi-view RGB images.
Figure~\ref{fig:pipeline} shows the overview. 

\subsection{Stage 1: Geometry Initialization}
\label{sec:2d_gaussian_splatting}
We initialize scene geometry using 2D Gaussian splatting~\cite{huang20242d} applied to multi-view RGB images, which has been shown to be robust under diverse ambient lighting conditions.
{Each Gaussian is parameterized by its geometric attributes, including its center, principal tangent directions, scaling factors , and opacity, as well as view-dependent radiance modeled as spherical harmonics. 

The final pixel is rendered with alpha blending as
\begin{equation}\label{eq:alpha_blending}
I = \sum_{i=1}^{M} T_{i} \alpha_{i} R_i,\quad
T_i = \prod_{j=1}^{i-1} (1 - \alpha_j),
\end{equation}
{where $M$ is the number of Gaussians splatted onto each rendered pixel.
$R_i$ is radiance of each Gaussian, $\alpha_i$ is computed by the multiplication of opacity and Gaussian influence between the Gaussian center and the ray.}  $T_i$ is the transmittance accumulated by the opacity of the top-most $j$ Gaussians. We optimize the radiance and geometric parameters of the Gaussians by minimizing the difference between the rendered and the measured RGB pixel values. Refer to the original paper for further details~\cite{huang20242d}.}
Note that the estimated radiance parameters are discarded for subsequent stages; only the geometric parameters are retained.

\subsection{Stage 2: NIR Flash Inverse Rendering}
\label{sec:nir_inverse_rendering}
\begin{figure}[t]
\centering
\includegraphics[width=\linewidth]{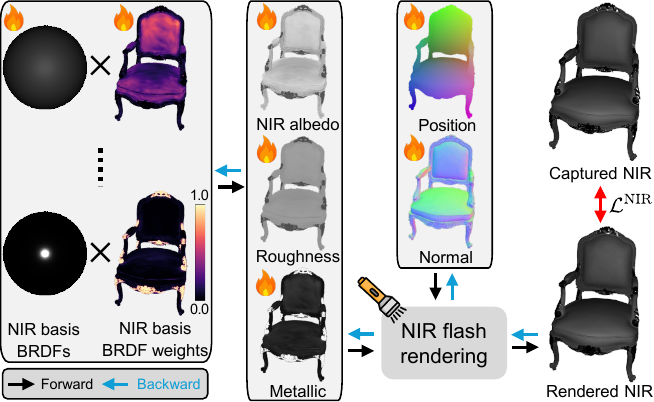}
\caption{\textbf{NIR flash inverse rendering.} We estimate NIR basis BRDFs and refine Gaussian geometry using multi-view NIR images with point-light shading, which is robust to ambient lighting. \NEW{The fire icon indicates trainable parameters.}}
\vspace{-5mm}
\label{fig:nir_pipeline}
\end{figure}

The second stage is the key to achieve robustness to ambient illumination.
Given the initialized Gaussians from Stage~1, this Stage~2 estimates NIR reflectance and refines geometry using multi-view NIR flash images $I^{\text{NIR}}$, which are dominated by point-light shading from the NIR flash.
Figure~\ref{fig:nir_pipeline} shows the overview.

We model the NIR point-light shading for each Gaussian $g$ as
\begin{equation}
R^{\text{NIR}}(g)=
L^{\text{NIR}}(\mathbf{i};g)
\,
f^{\text{NIR}}(\mathbf{i}, \mathbf{o};g)
\,
(\mathbf{i}\!\cdot\!\mathbf{n}),
\label{eq:NIR_image_formation}
\end{equation}
where $\mathbf{i}$ denotes the incident light direction from the NIR flash, $\mathbf{o}$ is the outgoing view direction, and $\mathbf{n}$ is the surface normal.
$L^{\text{NIR}}(\mathbf{i};g)$ is the NIR light intensity arriving at $g$.
The term $f^{\text{NIR}}(\mathbf{i}, \mathbf{o};g)$ denotes the BRDF of the Gaussian in the NIR spectrum.

Inspired by previous flash-based RGB BRDF estimation~\cite{chung2024differentiable, practical, chung2025differentiable}, we represent the NIR reflectance $f^{\text{NIR}}$ using a basis BRDF formulation, expressing it as a weighted summation of $N$ NIR basis BRDFs $\{f^{\text{NIR}}_{k}\}_{k=1}^{N}$:
\begin{equation}
f^{\text{NIR}}(\mathbf{i}, \mathbf{o};g)
=
\sum_{k=1}^{N} w_k(g) \, f^{\text{NIR}}_{k}(\mathbf{i}, \mathbf{o}),
\qquad
\sum_{k=1}^{N} w_k(g) = 1,
\label{eq:NIR_basis_brdf}
\end{equation}
where $w_k(g)$ denotes the mixture weight of the $k$-th basis BRDF for each Gaussian $g$.
Each basis BRDF is defined following the Disney BRDF model:
\begin{equation}
f^{\text{NIR}}_{k} \! (\mathbf{i}, \mathbf{o})
 \! = \! 
\frac{1  \! -  \!  m_k}{\pi}\,  \! \rho^{\text{NIR}}_{k}  
+ 
\frac{
D(\mathbf{h}; \sigma_k)\,
F(\mathbf{o}, \mathbf{h}; \sigma_k, m_k)\,
G(\mathbf{i}, \mathbf{o}, \mathbf{n}; \sigma_k)
}{
4(\mathbf{n}\!\cdot\!\mathbf{i})(\mathbf{n}\!\cdot\!\mathbf{o})
},
\end{equation}
where $\mathbf{h} = (\mathbf{i} + \mathbf{o}) / \|\mathbf{i} + \mathbf{o}\|$ is the half vector.
{$D$, $F$, and $G$ denote the microfacet normal distribution, Fresnel term, and geometric attenuation, respectively.}  
Each basis BRDF is parameterized by NIR diffuse albedo $\rho^{\text{NIR}}_{k}$, roughness $\sigma_k$, and metallic $m_k$.

{Using the NIR radiance of Gaussian from Equation~\eqref{eq:NIR_image_formation}, we render the NIR image $\hat{I}^{\text{NIR}}$ via differentiable rasterization of Equation~\eqref{eq:alpha_blending}.}
Using the rendered $\hat{I}^{\text{NIR}}$ and the captured NIR images $I^{\text{NIR}}$, we optimize the following objective:
\begin{equation}
\mathcal{L}^{\text{NIR}}
=
\mathcal{L}^{\text{NIR}}_{\text{rec}}
+ \lambda_{\text{geom}}\,\mathcal{L}_{\text{geom}}
+ \lambda_{\text{mask}}\,\mathcal{L}_{\text{mask}}
+ \lambda_{\text{smooth}}\,\mathcal{L}_{\text{smooth}},
\label{eq:loss_function_nir}
\end{equation}
{where $\lambda_{\text{geom}}, \lambda_{\text{mask}}$, and $\lambda_{\text{smooth}}$ are hyperparameters that weigh each loss component.}
$\mathcal{L}^{\text{NIR}}_{\text{rec}}$ enforces photometric consistency between $\hat{I}^{\text{NIR}}$ and $I^{\text{NIR}}$,
$\mathcal{L}_{\text{geom}}$ is the depth--normal consistency loss~\cite{huang20242d},
$\mathcal{L}_{\text{mask}}$ aligns rendered and observed object masks,
and $\mathcal{L}_{\text{smooth}}$ is an edge-aware reflectance smoothness prior~\cite{R3DG2023}.
We minimize this loss via backpropagation, jointly optimizing the NIR basis BRDF parameters
$\{\rho^{\text{NIR}}_k, \sigma_k, m_k\}$,
the per-Gaussian mixture weights $\{w_k\}$,
and the Gaussian geometry.
{Further implementation details are provided in the Supplemental Document.}

\paragraph{Cross-spectral Roughness and Metallic}
After optimization, the per-Gaussian roughness and metallic are computed as weighted averages of all bases:
\begin{equation}
\sigma = \sum_{k=1}^{N} w_k\,\sigma_k,
\qquad
m = \sum_{k=1}^{N} w_k\,m_k.
\label{eq:RGB_roughness_metallic}
\end{equation}
We share these roughness and metallic parameters, estimated under NIR point-light shading, to the RGB domain, as these parameters are related to micro-geometry and material-type properties that are largely wavelength-independent. 
This shared representation enables ambient-robust inverse rendering in the subsequent RGB stage.

\begin{figure}[t]
\centering
\includegraphics[width=\linewidth]{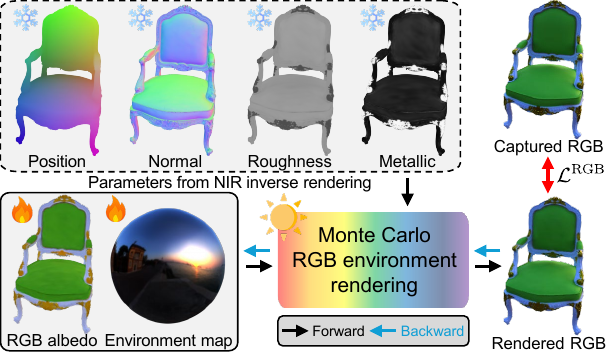}
\caption{\textbf{RGB environment inverse rendering.} Given the parameters from Stage~2, we estimate RGB albedo and environment map using multi-view RGB images. \NEW{The fire icon indicates trainable parameters, while the snowflake icon denotes fixed parameters.}}
\vspace{-5mm}
\label{fig:rgb_pipeline}
\end{figure}

\begin{table*}[t]
    \centering
    \caption{\textbf{\NEW{Quantitative evaluation on our synthetic and real-world dataset.}} }
    \resizebox{1.4\columnwidth}{!}{
    \begin{tabular}{c|rrr|r|r|r|r}
        \toprule[1pt]
            & \multicolumn{3}{c|}{RGB diffuse albedo} & Roughness & Normal & Relighting & Time \\
            & PSNR$\uparrow$ & SSIM$\uparrow$ & LPIPS$\downarrow$ & RMSE$\downarrow$ & MAE$\downarrow$ & PSNR$\uparrow$ & Hour$\downarrow$\\ \hline
            NeILF~\cite{yao2022neilf} & 16.61  & 0.8318 & 0.1409 & 0.4138 & 20.70 & N/A & 12\\
            TensoIR~\cite{tensoir}    & 17.40 & 0.9057 & 0.0875 & 0.1586 & 19.30 & 20.22 &6\\
            GS-IR~\cite{liang2024gs}  & 17.24 & 0.8812 & 0.1277 & 0.2207 & 16.37 & 27.17 &\textbf{0.7}\\
            R3DG~\cite{R3DG2023}      & 16.69 & 0.8871 & 0.1173 & 0.1371 & 13.35 & 26.11 &1\\
            IRGS~\cite{gu2024IRGS}    & 13.96 & 0.8580 & 0.0827 & 0.1174 & 10.48 & 26.45 &2\\ \hline
            WildLight~\cite{cheng2023wildlight} & 21.10 & 0.9328 & 0.1053 & 0.1472 & 7.84 & N/A  &6\\ \hline
            Ours                      & \textbf{27.89} & \textbf{0.9675} & \textbf{0.0680} & \textbf{0.0713} & \textbf{7.73} &\textbf{31.01} & 1.2\\
        \bottomrule[1pt]
    \end{tabular}}
    \label{tab:synthetic_comparison}
    \vspace{-3mm}
\end{table*}

\subsection{Stage 3: RGB Environment Inverse Rendering}
\label{sec:rgb_inverse_rendering}
At this Stage~3 shown in Figure~\ref{fig:rgb_pipeline}, the remaining unknowns are RGB diffuse albedo $\rho^{c}$ and RGB environment map $L_{\mathrm{env}}^{c}$ for $c \in \{R,G,B\}$. Accurate estimates of geometry, roughness, and metallic have been obtained in Stage~2.
From a set of 2D Gaussians with geometric and reflectance parameters, we rasterize RGB diffuse albedo, NIR diffuse albedo, roughness, metallic, position, and normal images.
We start by modeling the image formation of RGB images captured under ambient lighting.
The RGB intensity at pixel $p$ and color channel $c$ is given as
\begin{equation}
I^{c}(p)
=
\int_{\Omega}
L^{c}(\mathbf{i})\,
f^{c}(\mathbf{i}, \mathbf{o})\,
(\mathbf{i}\!\cdot\!\mathbf{n})
\, d\mathbf{i},
\label{eq:RGB_image_formation}
\end{equation}
where $L^{c}(\mathbf{i})$ denotes the incident hemispherical illumination, decomposed into direct and indirect components as
\begin{equation}
L^{c}(\mathbf{i})
=
V(\mathbf{i})\,L_{\mathrm{env}}^{c}(\mathbf{i})
+
L_{\mathrm{indirect}}^{c}(\mathbf{i}),
\end{equation}
where $L_{\mathrm{env}}^{c}$ represents the unknown direct environment lighting.
$V(\mathbf{i})$ is the visibility term and and $L_{\mathrm{indirect}}^{c}$ is the indirect illumination, which are both estimated using Gaussian ray tracing~\cite{gu2024IRGS}.

For the RGB BRDF $f^{c}$, we employ the Disney BRDF model:
\begin{align}\label{eq:rgb_brdf}
f^{c}(\mathbf{i}, \mathbf{o}) 
= \frac{1 - m}{\pi}\,\rho^{c}
 + \frac{D(\mathbf{h}; \sigma)\, F(\mathbf{o}, \mathbf{h}; \sigma, m)\, G(\mathbf{i}, \mathbf{o}, \mathbf{n}; \sigma)}
   {4\, (\mathbf{n}\!\cdot\!\mathbf{i})(\mathbf{n}\!\cdot\!\mathbf{o})},
\end{align}
where the diffuse albedo $\rho^{c}$ is the only unknown.
The roughness $\sigma$ and metallic $m$ are known from Stage~2.
Hence, we do not employ basis BRDFs at this stage, avoiding loss in high-frequency details of the reconstructed RGB diffuse albedo.

For computational efficiency, we evaluate Equation~\eqref{eq:RGB_image_formation} using physics-based Monte Carlo integration with multiple importance sampling (MIS)~\cite{veach1995optimally}:
\begin{equation}
I^{c}(p)
\approx
\sum_{s \in \{b,l\}}
\frac{1}{N_{s}}
\sum_{j=1}^{N_s}
w_{s}(\mathbf{i}^{j}_{s})
\frac{
L^c(\mathbf{i}^{j}_{s})
f^{c}(\mathbf{i}^{j}_{s}, \mathbf{o})
(\mathbf{i}^{j}_{s}\!\cdot\!\mathbf{n})
}{
p_{s}(\mathbf{i}^{j}_{s})
},
\label{eq:RGB_image_formation_MC}
\end{equation}
where $N_{s}$ is the number of samples for each sampling strategy.
The BRDF sampling distribution $p_b$ follows a cosine-weighted GGX distribution~\cite{cook1982reflectance} and the light sampling distribution $p_l$ is proportional to the illumination $L^{c}_{env}$.
The balance heuristic weight is defined as
\begin{equation}
w_{s}(\mathbf{i}_{s})
=
\frac{N_s p_s(\mathbf{i}_{s})}{N_b p_b(\mathbf{i}_{s}) + N_l p_l(\mathbf{i}_{s})}.
\label{eq:balance_heuristic}
\end{equation}

Using Equation~\eqref{eq:RGB_image_formation_MC}, we render RGB images $\hat{I}^{c}$ and compare them with the captured images $I^{c}$ to jointly optimize the RGB diffuse albedo $\rho^{c}$ and the RGB environment map $L_{\mathrm{env}}^{c}$ by minimizing the following loss:
\begin{equation}\label{eq:loss_function_rgb}
\mathcal{L}^{\text{RGB}}
=
\mathcal{L}^{\text{RGB}}_{\text{rec}}
+
\lambda_{\text{RGB-edge}}\,
\mathcal{L}_{\text{RGB-edge}},
\end{equation}
{where $\lambda_{\text{RGB-edge}}$ is balancing weight.}
$\mathcal{L}^{\text{RGB}}_{\text{rec}}$ is the RGB reconstruction loss between the rendered image and the captured image. 
$\mathcal{L}_{\text{RGB-edge}}$ is the regularizer for the RGB diffuse albedo image defined as 
\begin{equation}\label{eq:loss_rgb_smooth}
\mathcal{L}_{\text{RGB-edge}}
=
\exp\!\big(-k\,|\nabla \rho^{\text{NIR}}|\big)
\,|\nabla \rho^{\text{RGB}}|,
\end{equation}
{where $k$ is a hyperparameter to weigh the edge of NIR albedo $\rho^{\text{NIR}}$} image.
This term prevents shadows from being baked into the estimated RGB albedo by using the shadow-free NIR albedo as guidance.  
\section{Assessment}
\label{sec:results}


\subsection{Validation of the RGB--NIR BRDF Model}
\begin{figure}[t]
\centering
\includegraphics[width=\linewidth]{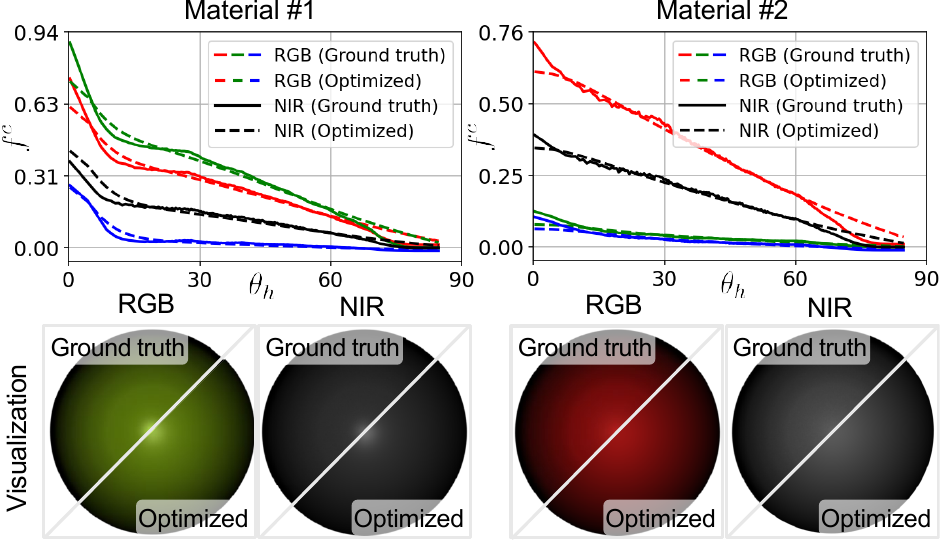}
\caption{\textbf{Validation of the RGB--NIR BRDF model.}
Sharing roughness and metallic across RGB and NIR channels enables accurate modeling of the measured hyperspectral BRDFs~\cite{dupuy2018adaptive}.}
\vspace{-7mm}
\label{fig:BRDF_analysis}
\end{figure}

Our RGB--NIR inverse rendering framework shares roughness and metallic across the R, G, B, and NIR channels, while estimating diffuse albedo independently for each channel.
We validate this assumption by fitting the roughness $\sigma$, metallic $m$, and per-channel diffuse albedo $\rho^{c}$ to the hyperspectral BRDF dataset~\cite{dupuy2018adaptive}.
Figure~\ref{fig:BRDF_analysis} shows that the fitted RGB--NIR BRDFs closely match the measurements, supporting our assumption.
Detailed quantitative evaluation is provided in the Supplemental Document.

\subsection{Ambient-robust Inverse Rendering}

\begin{figure*}[t]
\centering
\includegraphics[width=\linewidth]{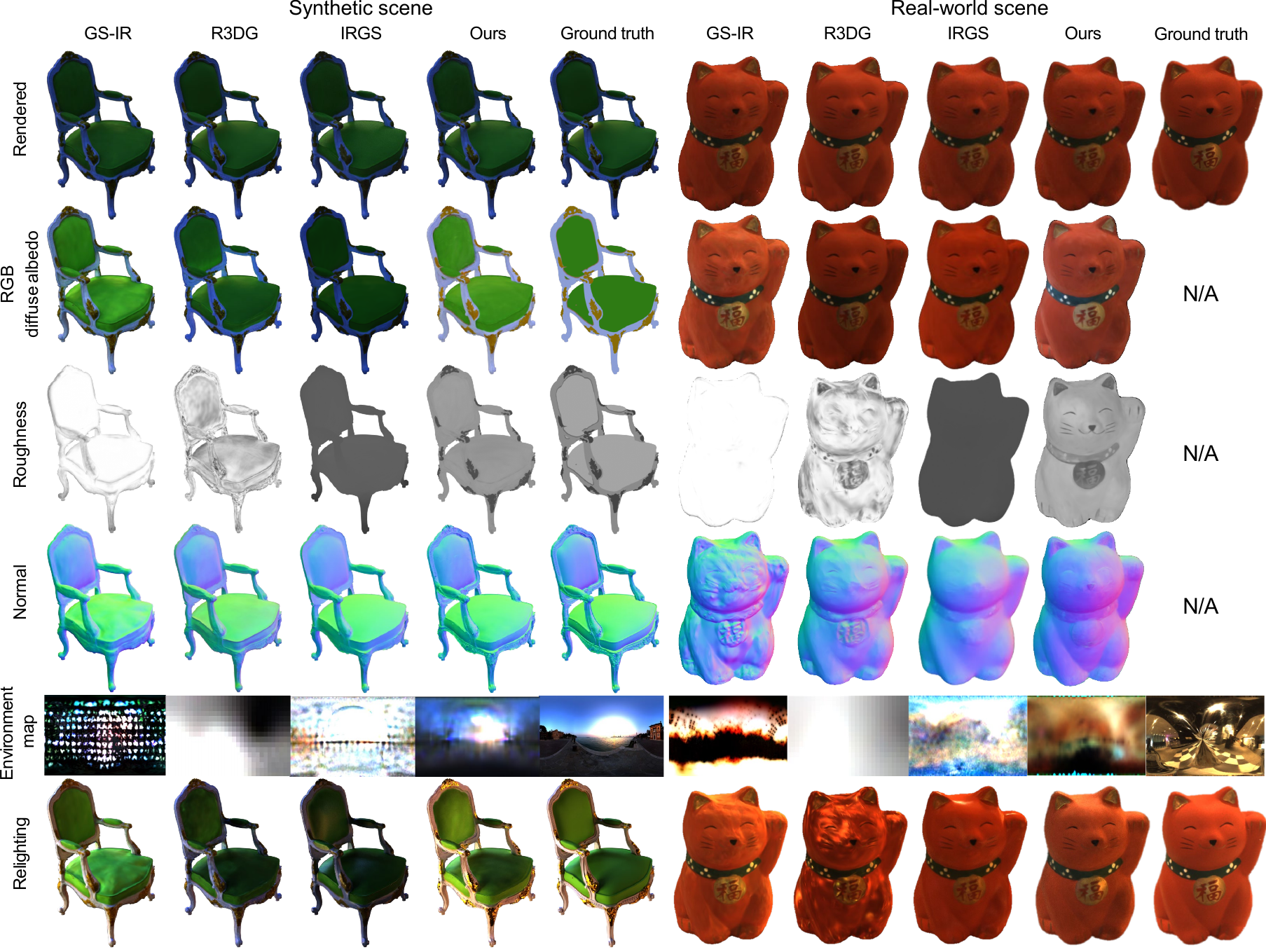}
\caption{\textbf{Comparison with passive RGB inverse rendering methods.} 
Our method enables ambient-robust reconstruction, outperforming passive RGB inverse rendering approaches: R3DG~\cite{R3DG2023}, GS-IR~\cite{liang2024gs}, and IRGS~\cite{gu2024IRGS}. 
\vspace{-4mm}}
\label{fig:synth_results}
\end{figure*}

\begin{figure}[t]
\centering
\includegraphics[width=0.9\linewidth]{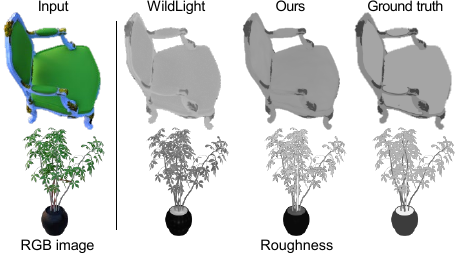}
\caption{\textbf{Comparison with active RGB inverse rendering.} 
\NEW{Our method reconstructs more accurate reflectance than WildLight~\cite{cheng2023wildlight}.}
}
\label{fig:synth_wild}
\vspace{-8mm}
\end{figure}

\begin{figure}[t]
\centering
\includegraphics[width=\linewidth]{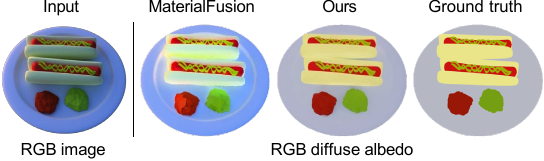}
\caption{\textbf{Comparison with diffusion-based inverse rendering.} 
We reconstruct RGB diffuse albedo more accurately than MaterialFusion~\cite{litman2025materialfusion}, with less shading contamination and color bias.}
\label{fig:diffusion_comparison}
\vspace{-6mm}
\end{figure}

\paragraph{Comparison}
\NEW{We compare our RGB–NIR inverse rendering method with passive RGB-based approaches, including NeILF~\cite{yao2022neilf}, TensoIR~\cite{tensoir}, R3DG~\cite{R3DG2023}, GS-IR~\cite{liang2024gs}, and IRGS~\cite{gu2024IRGS}.
Table~\ref{tab:synthetic_comparison} shows that our method achieves high accuracy in geometry and reflectance estimation across the entire synthetic dataset under diverse environment maps.
Using PBR assets estimated under environment map 1, we perform relighting under environment maps 2–4 for both synthetic and real-world scenes, where our method outperforms all baselines. Note that NeILF and WildLight do not support relighting under novel illumination.
As shown in Figure~\ref{fig:synth_results}, our approach enables qualitatively accurate and robust reconstruction across both synthetic and real-world scenarios.}
This is because passive RGB inverse rendering struggles to disentangle ambient illumination and surface reflectance, whereas our RGB-NIR method achieves accurate reconstruction using NIR point-light shading.
Next, we compare our method with the active RGB-based inverse rendering method, WildLight~\cite{cheng2023wildlight}.
WildLight uses RGB images captured under ambient lighting and ambient lighting with RGB flash.
Figure~\ref{fig:synth_wild} shows that WildLight results in degraded reflectance due to the ambiguity in separating RGB flash from RGB ambient illumination.
Last, we compare our method with the recent diffusion-based inverse rendering method, MaterialFusion~\cite{litman2025materialfusion}.
Figure~\ref{fig:diffusion_comparison} shows that MaterialFusion suffers from residual shading artifacts and color bias in the estimated diffuse albedo. Additional results are provided in the Supplementary Document.

\begin{figure*}[t]
\centering
\vspace{-2mm}
\includegraphics[width=\linewidth]{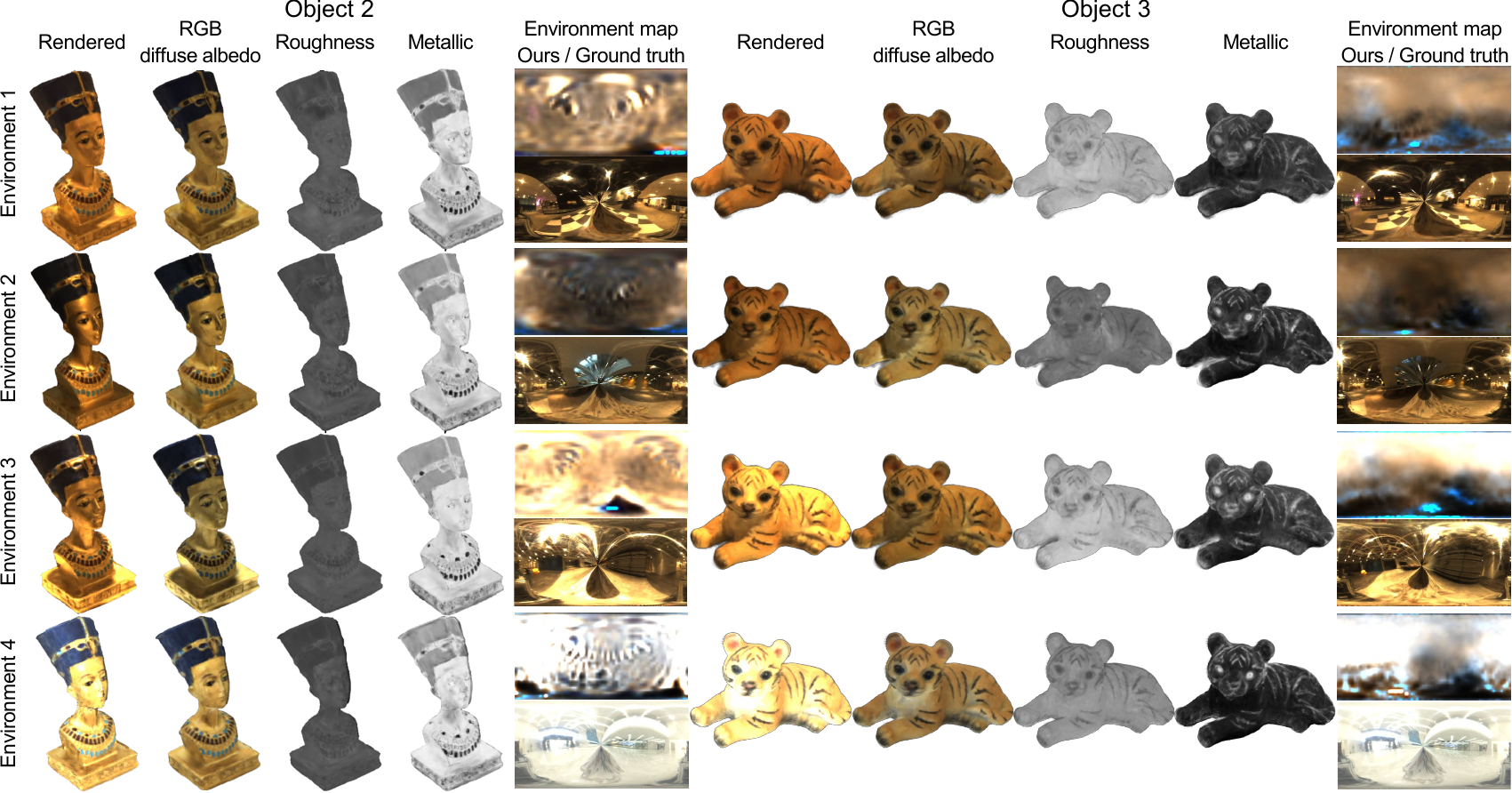}
\caption{\textbf{Ambient-robust reconstruction across environment maps on real-world dataset.}
{Our method reconstructs surface reflectance for real-world objects under multiple ambient illumination conditions, producing stable reflectance and environment estimation despite lighting variations.}}
\label{fig:real_consistency}
\vspace{-5mm}
\end{figure*}

\begin{figure}[t]
\centering
\includegraphics[width=\linewidth]{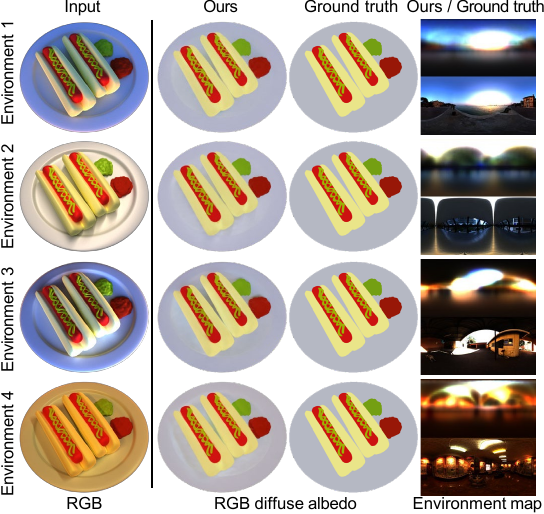}
\caption{\textbf{RGB-diffuse albedo reconstruction across environment maps.}
{Across diverse environment maps, our method robustly recovers consistent RGB diffuse albedo for synthetic objects, demonstrating effective disentanglement of surface reflectance from varying ambient illumination}}
\label{fig:synthetic_consistency}
\vspace{-3mm}
\end{figure}

\begin{figure}[t]
\centering
\includegraphics[width=\linewidth]{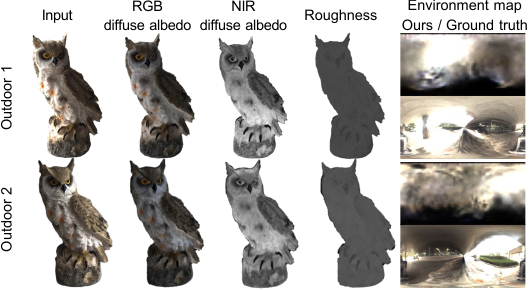}
\caption{\textbf{Reflectance reconstruction under real-world outdoor illuminations} 
\NEW{Our method reconstructs consistent reflectance under two different outdoor illuminations containing NIR ambient light.}
}
\vspace{-3mm}
\label{fig:ablation_env}
\end{figure}

\paragraph{Ambient Robustness}
Figure~\ref{fig:real_consistency} shows that our RGB--NIR inverse rendering enables consistent reconstructions of reflectance and environment map under four different indoor environment maps on real-world dataset. 
Figure~\ref{fig:synthetic_consistency} demonstrates ambient-robust reconstruction of RGB diffuse albedo for synthetic dataset under different ambient illuminations. \NEW{Figure~\ref{fig:ablation_env} extends this evaluation to outdoor settings, showing reliable recovery of RGB and NIR diffuse albedo as well as roughness under illuminations with NIR ambient light.}

\vspace{-1mm}

\begin{figure}[t]
\centering
\includegraphics[width=\linewidth]{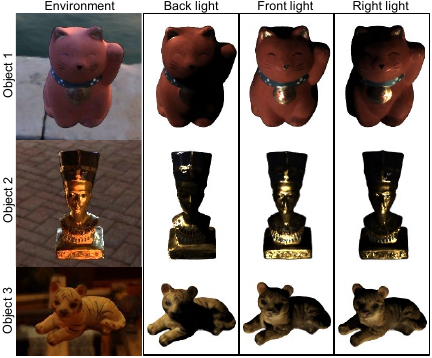}
\caption{\textbf{Relighting results for real-world objects}. Our method enables realistic relighting of real-world objects for diverse illumination setups.}
\label{fig:real_relighting}
\vspace{-3mm}
\end{figure}

\paragraph{Realistic Relighting}
While the reconstructed geometric and reflectance parameters can be used for many applications, herein demonstrates realistic relighting as an example.
Figure~\ref{fig:real_relighting} shows the relit results under both environment maps and multiple point light configurations, demonstrating natural appearance such as specular highlights and shading. See the Supplemental Video for additional relighting results.

\subsection{Ablation Study}

\begin{figure}[t]
\centering
\includegraphics[width=\linewidth]{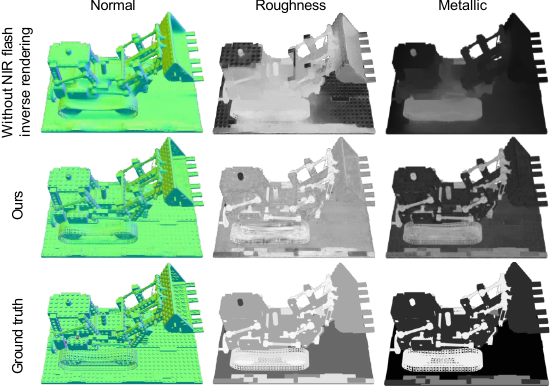}
\caption{\textbf{Impact of NIR flash inverse rendering.} 
NIR flash inverse rendering improves reconstruction accuracy of our method by leveraging NIR point-light shading.}
\label{fig:ablation_nir}
\vspace{-1mm}
\end{figure}

\paragraph{Importance of NIR Inverse Rendering}
We evaluate the contribution of the NIR flash inverse rendering stage through an ablation study that excludes Stage~2 (NIR flash inverse rendering), using only Stage~1 and Stage~3 of the pipeline.
As shown in Figure~\ref{fig:ablation_nir}, incorporating NIR inverse rendering substantially improves reconstruction accuracy.
This improvement arises because NIR flash provides a controlled illumination setting that enables reliable recovery of geometry and NIR reflectance parameters, which in turn stabilize the subsequent RGB inverse rendering stage.

\begin{figure*}[t]
\centering
\includegraphics[width=\linewidth]{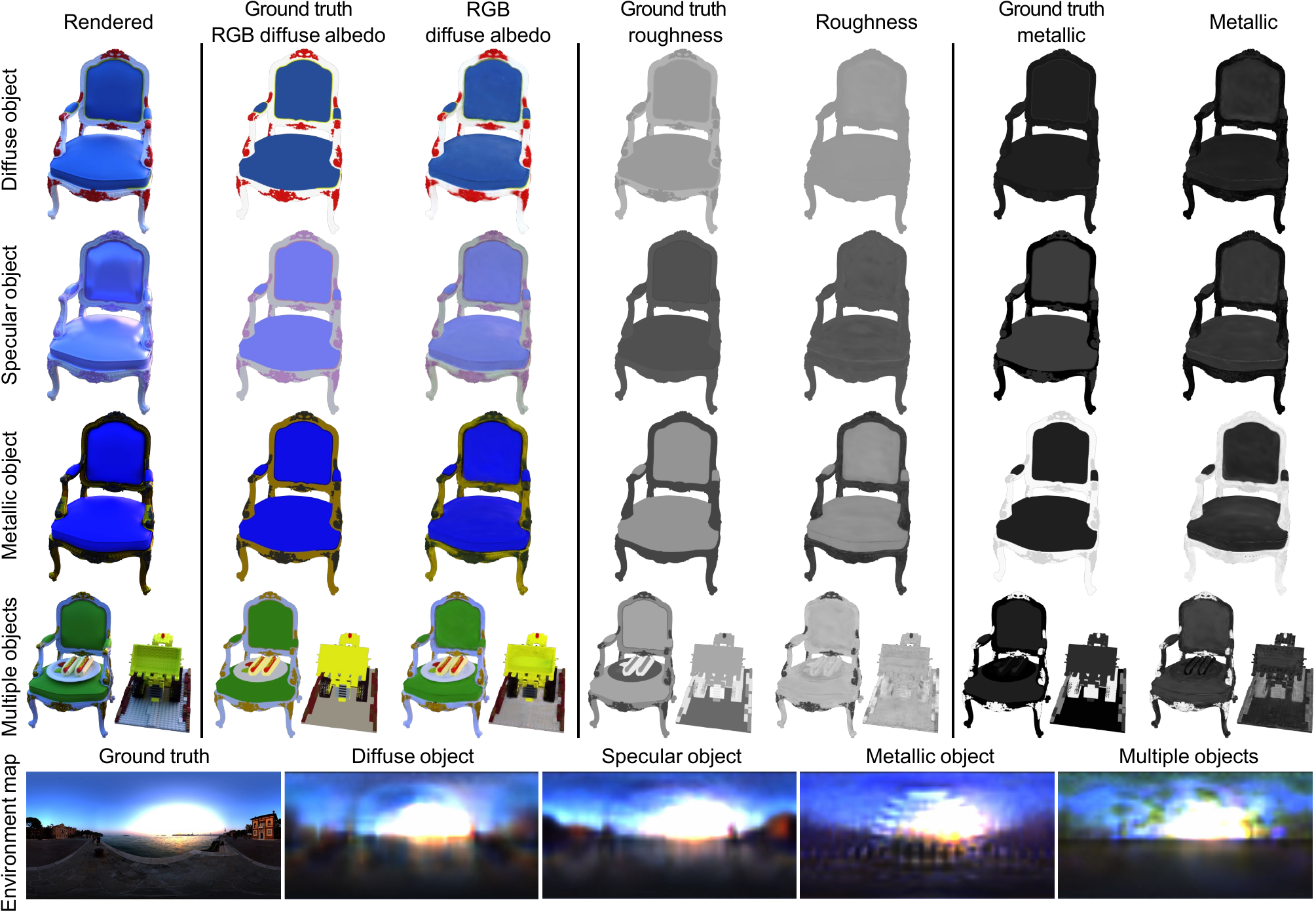}
\caption{\textbf{Dependency on material types and scene complexity.} 
Our method is robust across diverse material appearances, enabling accurate reconstruction for diffuse, specular, and metallic surfaces. We also reconstruct accurate reflectance and ambient illumination of multiple objects.}
\label{fig:material_comparison}
\vspace{-4mm}
\end{figure*}

\paragraph{Dependency on Material Types}
We evaluate the impact of object materials on the performance of our RGB--NIR inverse rendering.
To this end, we modify the chair object in our dataset to create three scenes with distinct material appearances—diffuse, specular, and metallic—under the same environment map.
As shown in Figure~\ref{fig:material_comparison}, our method achieves accurate reconstruction across all three material types.
For specular materials, we recover more accurate environment map that preserves both low- and high-frequency illumination details compared to the diffuse and metallic cases.
We attribute this behavior to the stronger and more localized illumination cues provided by specular reflections.

\paragraph{Dependency on Scene Complexity}
We evaluate the impact of scene complexity on the performance of our method by increasing the number of objects in a scene. 
We compose three objects (chair, hotdog, lego) into a scene and assess reconstruction quality. As shown in Figure~\ref{fig:material_comparison}, our method reconstructs accurate surface reflectance and environment map across all objects in the scene.

\subsection{Limitations}
\NEW{We acknowledge several limitations of our approach. Scaling to meter-scale scenes remains challenging due to mask extraction errors, complex indirect illumination, occlusions, and spatially-varying environment maps. Our method also struggles with extreme mirror-like materials due to imperfect geometry initialization and unreliable NIR shading cues caused by self-reflections. In addition, strong ambient NIR illumination, such as direct sunlight, significantly reduces the effective dynamic range of NIR-flash signals, leading to degraded performance. This issue may be mitigated by using a brighter NIR source or reducing the object-to-sensor distance. Finally, some materials such as dyes and leaves may exhibit inconsistent reflectance properties between RGB and NIR domains, potentially violating our shared-parameter assumption.}
\section{Conclusion}
\label{sec:conclusion}
We have introduced an ambient-robust inverse rendering method using \NEW{active} RGB–NIR imaging, addressing the challenge of recovering geometry and reflectance under uncontrolled ambient illumination. Our method leverages NIR illumination, which is imperceptible to human observers, to obtain stable point-light shading invariant to ambient lighting conditions. By combining multi-view ambient RGB images with flash-based NIR observations, we enable accurate reconstruction of object geometry and reflectance across diverse lighting environments.
With our RGB--NIR BRDF model that shares roughness and metallic across RGB and NIR channels while estimating per-channel diffuse albedo, we achieve physically consistent reconstruction across spectra. Our \NEW{active} RGB–NIR imaging system enables collecting the first multi-view RGB–NIR inverse rendering dataset captured under multiple ambient illumination conditions. We believe this work serves as a step towards practical, automatic, and robust scene reconstruction.

\begin{acks}
This work was supported by the National Research Foundation of Korea (NRF) grant funded by the Korea government (MSIT) (RS-2024-00438532, No. RS-2023-00211658), the IITP (Institute of Information \& Communications Technology Planning \& Evaluation)-ITRC (Information Technology Research Center) grant funded by the Korea government (MSIT) (IITP-2026-RS-2024-00437866), and the Institute of Information \& communications Technology Planning \& Evaluation (IITP) grant funded by the Korea government (MSIT) (No. RS-2024-0045788).
\end{acks}

\bibliographystyle{ACM-Reference-Format}
\bibliography{references}

\clearpage

\end{document}